\documentclass[10pt,twocolumn,twoside]{IEEEtran} 
\usepackage{authblk}

\usepackage{epsfig}
\usepackage{booktabs}
\usepackage{graphicx}
\usepackage[format=hang,font={sf,scriptsize},labelfont=bf]{subfig}
\usepackage[font=small,skip=0pt]{caption}
\usepackage{cleveref}
\usepackage{chngcntr}

\usepackage{diagbox}
\usepackage{color}

\usepackage[thmmarks]{ntheorem}
{
	\theoremstyle{nonumberplain}	
	\theoremheaderfont{\bfseries}
	\theorembodyfont{\normalfont}
	\theoremsymbol{\mbox{$\Box$}}

}

\let\labelindent\relax
\usepackage{enumitem}
\newlist{myEnumerate}{enumerate}{2}
\setlist[myEnumerate,1]{leftmargin=*,topsep=0pt,itemsep=0pt,parsep=0pt,label=\normalfont\textbf{\arabic*)}}
\setlist[myEnumerate,2]{leftmargin=*,topsep=0pt,itemsep=0pt,parsep=0pt,label=\normalfont(\alph*)}

\usepackage{algorithm,algorithmic}
\newlength\myindent
\usepackage{xspace}

\usepackage{amsmath,amsfonts,amssymb,bm,yhmath,accents}
\usepackage{mathtools}
\usepackage{multirow}

\newcommand{\etal}{\textit{et al}.}

\usepackage[pagebackref=true,breaklinks=true,letterpaper=true,colorlinks,bookmarks=false]{hyperref}
\usepackage{soul}

\begin{document}

\title{Learning Depth from Single Images with Deep Neural Network Embedding Focal Length}

\author{Lei~He,~Guanghui~Wang (\textit{Senior Member}, IEEE)~and~Zhanyi~Hu 
\IEEEcompsocitemizethanks{\IEEEcompsocthanksitem L. He and Z. Hu are with University of Chinese Academy of Sciences, Beijing 100049, China and National Laboratory of Pattern Recognition, Institute of Automation, Chinese Academy of Sciences, Beijing 100190, China. Z. Hu is also with CAS Center for Excellence in Brain Science and Intelligence Technology, Chinese Academy of Sciences, Beijing 100190, China. \protect\\
E-mail: lei.he@nlpr.ia.ac.cn; huzy@nlpr.ia.ac.cn.}
\IEEEcompsocitemizethanks{\IEEEcompsocthanksitem G. Wang is with the Department of Electrical Engineering and Computer Science, University of Kansas, Lawrence, KS 66045, USA. \protect\\
E-mail:  ghwang@ku.edu.}
}

\maketitle

\thispagestyle{empty}

\begin{abstract}
Learning depth from a single image, as an important issue in scene understanding, has attracted a lot of attention in the past decade. The accuracy of the depth estimation has been improved from conditional Markov random fields, non-parametric methods, to deep convolutional neural networks most recently. However, there exist inherent ambiguities in recovering 3D from a single 2D image. In this paper, we first prove the ambiguity between the focal length and monocular depth learning, and verify the result using experiments, showing that the focal length has a great influence on accurate depth recovery. In order to learn monocular depth by embedding the focal length, we propose a method to generate synthetic varying-focal-length dataset from fixed-focal-length datasets, and a simple and effective method is implemented to fill the holes in the newly generated images. For the sake of accurate depth recovery, we propose a novel deep neural network to infer depth through effectively fusing the middle-level information on the fixed-focal-length dataset, which outperforms the state-of-the-art methods built on pre-trained VGG. Furthermore, the newly generated varying-focal-length dataset is taken as input to the proposed network in both learning and inference phases. Extensive experiments on the fixed- and varying-focal-length datasets demonstrate that the learned monocular depth with embedded focal length is significantly improved compared to that without embedding the focal length information.

\end{abstract}


\section{Introduction}
Scene depth inference from a single image is currently an important issue in machine learning~\cite{saxena2009make3d,karsch2012depth,eigen2015predicting,laina2016deeper,xu2017multi}. The underlying rationale of this problem is the possibility of human depth perception from single images. The task here is to assign a depth value to every single pixel in the image, which can be considered as a dense regression problem. Depth information can benefit many challenging computer vision problems, such as semantic segmentation~\cite{hazirbas2016fusenet,Cao2016Exploiting}, pose estimation~\cite{Shotton2013Efficient}, and object detection~\cite{Song2016Deep}.

During the past decade, significant effort has been made in the research community to improve the performance of monocular depth learning, and significant accuracy has been achieved thanks to the rapid development and advances of deep neural networks. However, most available methods overlook one key problem: the ambiguity between the scene depth and the camera's focal length. Because the 3D-to-2D object imaging process must satisfy some strict projective geometric relationship, however, without prior knowledge on the camera's intrinsic parameters, it is impossible to infer the true depth from a single image.

In this paper, in order to remove the ambiguity caused by the unknown focal length, we propose a novel deep neural network to learn the monocular depth by embedding the focal length information. However, the datasets used in most machine learning methods are all of fixed-focal-length, such as the NYU dataset~\cite{Silberman:ECCV12}, the Make3D dataset~\cite{saxena2009make3d}, and the KITTI dataset~\cite{geiger2013vision}. To prepare for learning monocular depth with focal length, datasets with varying focal lengths are required so that the camera¡¯s intrinsic information should be taken into account in both the learning and the inference phases. However, considering the labor in building a new varying-focal-length dataset, it is desirable to transform the existing fixed-focal-length datasets into those of varying-focal-length. we first introduce a method to generate varying-focal-length dataset from fixed-focal-length dataset, like Make3D and NYU v2, and a simple and effective method is proposed to fill the holes produced during the image generation. The transformed datasets are demonstrated to make great contribution in depth estimation.

In order to learn fine-grained monocular depth with focal length, we propose an effective neural network to predict accurate depth, which achieves competitive performance as compared with the state-of-the-art methods, and further embedding the focal length information into the proposed model. In the literature, almost all works for pixel-wise prediction exploit an Encoder-Decoder network~\cite{badrinarayanan2015segnet,noh2015learning} to infer the labels of pixels. To predict accurate labels, two general attempts have been made to address the problem. One is to integrate middle layer features~\cite{long2015fully,hariharan2015hypercolumns,badrinarayanan2015segnet,ronneberger2015u,lin2016refinenet}, the other is to effectively exploit the multi-scale information and the decoder side outputs~\cite{eigen2015predicting,xu2017multi,liu2016richer,xie2015holistically}. Inspired by the idea of fusing the middle-level information, we propose a novel end-to-end neural network to learn fine-grained depth from single images with embedded focal length. The proposed network is composed of four parts: the first part is built on the pre-trained VGG models, followed by the global transformation layer and upsampling architecture to produce depth with high resolution, the third part effectively integrates the middle-level information to infer the structure details, converting the middle-level information to the space of the depth, and the last part embeds the focal length into the global information.

The proposed network is extensively evaluated on the Make3D, NYU v2, and KITTI datasets. We first perform the experiments without the embedded focal length, and better performance than the state-of-the-art techniques is achieved in both quantitative and qualitative terms. Then, it is further evaluated with the embedded focal length on the newly generated varying-focal-length datasets for comparison. The experimental results show that depths inferred from the model with embedded focal length significantly outperform those without the focal length in all error measures, it also demonstrates that the focal length information is very useful for the depth extraction from a single image.

In summary, the contributions of this paper are four-fold. First, we prove that the ambiguity between the focal length and the depth estimation from a single image, and further demonstrate the result using real images. Second, we propose a method to generate varying-focal-length images from fixed-focal-length images, which are visually plausible. Third, based on the classical Encoder-Decoder network, a novel neural network model is proposed to learn the fine-grained depth from single images, by virtue of effectively fusing the middle-level information. Finally, given the newly generated varying-focal-length datasets, we revise the fine-grained network by embedding the focal length information. The experimental evaluation shows that the depth inference with known focal length achieves significantly better performance than the one without the focal length information. The source code and the generated datasets will be available on the author¡¯s website.

The rest of this paper is organized as follows: Section~\ref{sec:related_work} introduces the related works. The ambiguity between the focal length and monocular depth estimation is discussed in Section~\ref{sec:ambiguity}. Section~\ref{sec:data_transformation} describes the generating process from fixed-focal-length dataset to varying-focal-length dataset. The proposed fine-grained network embedding focal length information is elaborated in Section~\ref{sec:network}, and the experimental results on the four datasets are reported in Section~\ref{sec:experiments}. The paper is concluded in Section~\ref{sec:conclusion}.

\section{Related Work}\label{sec:related_work}
Depth extraction from single images has received a lot of attention in recent years, while it remains a very hard problem due to the inherent ambiguity. To tackle this problem, classic methods~\cite{hoiem2005automatic,schwing2012efficient,hedau2010thinking,saxena2005learning,saxena20083,wang2009can,wang2005single,wang2005camera} usually make strong geometric assumptions that the scene structure consists of horizontal planes, vertical walls and superpixels, employing the Markov random field (MRF) to infer the depth by leveraging the handcrafted features. One of the first work, proposed by Hoiem \etal~\cite{hoiem2005automatic}, creates realistic-looking reconstructions of outdoor images by assuming planar scene composition. In~\cite{schwing2012efficient,hedau2010thinking}, simple geometric assumptions have proven to be effective in estimating the layout of a room. In order to improve the accuracy of the depth-based methods, Saxena \etal~\cite{saxena2005learning,saxena20083} utilized MRF to infer depth from both local and global features extracted from the image. In addition, superpixels~\cite{achanta2012slic} are introduced in the MRF formulation to enforce neighboring constraints. The work has also been extended to 3D reconstruction of scenes~\cite{saxena2009make3d}.

Non-parameter algorithms~\cite{karsch2012depth,liu2011sift,konrad20122d,liu2014discrete} are another kind of classical methods for learning the depth from a single image, relying on the assumption that the similarities between regions in the RGB images imply similar depth cues as well. After clustering the training dataset based on the global features (e.g. GIST~\cite{oliva2001modeling}, HOG~\cite{dalal2005histograms}), these methods first search the candidate RGB-D of the input RGB image in the feature space, then, the candidate pairs are warped and fused to obtain the final depth. Karsch~\etal~\cite{karsch2012depth} proposed a depth transfer method to warp the retrieved RGB-D using SIFT flow~\cite{liu2011sift}, followed by a global optimization framework to smooth the resulting depth. He~\etal~\cite{he2016fast} employed a sparse SIFT flow to speed up the depth inference based on the work~\cite{karsch2012depth}. Konrad \etal~\cite{konrad20122d} computed a median over the retrieved depth maps followed by cross-bilateral filtering for smoothing. Instead of warping the retrieved candidates, Liu \etal~\cite{liu2014discrete} explored continuous variables to represent the depth of image superpixels and discrete ones to encode relationships between neighboring superpixels, formulating the depth estimation as an optimization problem of the discrete-continuous graphical model. For learning the indoor depth, Zhuo \etal~\cite{zhuo2015indoor} exploited the structure of the scene at different levels to learn depth from a single image.

Recently, convolutional neural networks have seen remarkable advances in the high-level problems of computer vision, which have also been applied with great success to depth extraction from single images~\cite{eigen2014depth,eigen2015predicting,liu2015deep,li2015depth,wang2015towards,roy2016monocular,laina2016deeper,xu2017multi}. There exist two major approaches for the task of depth estimation in the related references: multi-scale training technique and super-pixel pooling with conditional random field (CRF) algorithm. In order to accelerate the convergence of the parameters during the training phase, most of the works are built upon winning architectures of the ImageNet Large Scale Visual Recognition Challenge (ILSVRC)~\cite{russakovsky2015imagenet}, often initializing their networks with Alex~\cite{krizhevsky2012imagenet}, VGG~\cite{simonyan2014very}, or ResNet~\cite{he2016deep}. Eigen \etal~\cite{eigen2014depth} first addressed this issue by fusing the depths from the global network and refined network. Their work later was extended to use a multi-scale convolutional network to predict depth, normal and semantic label from a single image in a deeper neural network~\cite{eigen2015predicting}. Other methods to obtain the fine-grained depth leveraged the representation of the neural network and the inference of the CRFs. Liu \etal~\cite{liu2015deep} presented a deep convolutional neural field model based on fully convolutional networks and a novel superpixel pooling method, combining the strength of deep CNN and the continuous CRF into a unified CNNs framework. Li \etal~\cite{li2015depth} and Wang \etal~\cite{wang2015towards} leveraged the benefit of the hierarchical CRFs to refine their patch-wise predictions from superpixel down to pixel level. Roy \etal~\cite{roy2016monocular} combined random forests and convolutional
neural networks to tackle the depth estimation. Laina \etal~\cite{laina2016deeper} built a neural network on ResNet, followed by designed up-sampling blocks to obtain high resolution depth. However, the middle-level features are not fused into the network to obtain detailed information of the depth. Based on the multi-scale network~\cite{eigen2014depth,eigen2015predicting}, Dan~\etal~\cite{xu2017multi} effectively exploited the side outputs of deep networks to infer depth by reformulating the continuous CRFs of the monocular depth estimation as sequential deep networks.

For all these depth learning methods, the experimental datasets are usually created by Kinect or laser scanner, where the RGB camera has a fixed focal length. In other words, currently the available depth datasets in the literature are all of fixed-focal-length. However, there exists an inherent ambiguity between monocular depth estimation and focal length, as described in our work~\cite{he46inherent}. Without knowing the camera's focal length, the depth can not be truly estimated from a single image. In order to remove the ambiguity, the camera's focal length should be considered in both depth learning and inference phases. In the following section, we will discuss the inherent ambiguity in depth recovery in detail.

\begin{figure}[t]
    \begin{center}
        \includegraphics[width=1\linewidth]{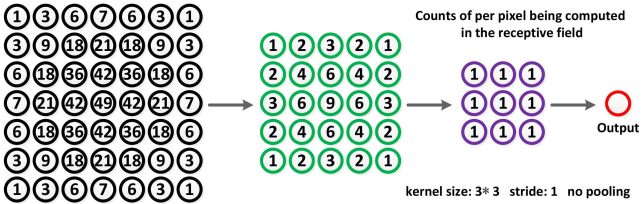}
    \end{center}
\caption{A novel method to visualize the receptive field. The number in the node represents the counts of per pixel being computed in the receptive field, which reveals that the receptive field obeys the Gaussian distribution and has a smaller size compared with the theoretical receptive field.}
\label{fig:receptive_field}
\end{figure}

\section{Inherent Ambiguity}\label{sec:ambiguity}
Scene depth refers to the distance from the camera optical center to the object along the optical axis. In deep learning based methods for the monocular depth estimation, the depth of each pixel is inferred by fusing global and contextual information, extracted from the corresponding receptive fields in the input image, followed by affine transformations and non-linear operations, as illustrated by the following equation.
\begin{equation}\label{eq:receptive_field}
    D_{i}=\sigma_{n}(w_{n}(\cdots\sigma_{1}(w_{1}x_{RF\_i}+b_{1})\cdots)+b_{n})
\end{equation}
where $ D_{i} $ is the depth of the pixel $ i $, $ x_{RF\_i} $ is the receptive field corresponding to the pixel $ i $ in the depth map, $ \sigma $ is the activation function and $ w, b$ are the parameters of the models.

In order to extract long range global information, the deep neural networks were introduced in the research community for monocular depth estimation. However, the deeper networks are very hard to train due to the vanishing gradient or exploding gradient. In addition, it may lead to another problem about the receptive fields. Note that for a specific network architecture, we can infer the theoretical receptive field associated with the output node in every layer. However, the contribution of various regions in the theoretical receptive field is not the same. To explore the role of each pixel location in the view-of-field, we propose a novel method to visualize the effective receptive field, as shown in Figure~\ref{fig:receptive_field}. From the output layer to the input layer, the counts of per pixel evolved in the convolution operation is obtained layer by layer.

In current nets of depth estimation from single images, the convolution operation usually adopts the technique of sharing weights in each channel, and the weights are initialized by sampling a Gaussian with zero mean and 0.01 variance. Once the network is trained, the parameters within each channel are fixed and shared. In addition, the number of use of each pixel for the final prediction could describe the complexity of the combination of network weights at each layer, including affine transformation and non-linear operation. The higher complexity of the combination, the better ability to character the problem of the corresponding task. In a statistical sense, this number represents that the pixel information is frequently used in monocular depth estimation, regardless of the weights, which makes it able to view the contribution of each pixel. It is observed that the deeper the depth of the network, the larger the value in the middle of the receptive field, while the one along the edge is in the opposite, which reveals that the actual receptive field is smaller than the theoretical receptive field, and it also obeys the Gaussian distribution as described in Luo~\etal~\cite{luo2016understanding}. In order to enlarge the view-of-field in the specific network, a fully connected layer is a better choice when the resolution of the feature maps is very small.

The methods for monocular depth estimation are based on the assumption that the similarities between regions in the RGB images imply similar depth cues as well. There exists an inherent ambiguity between the focal length and the scene depth learned from a single image, as analyzed in the following.

\begin{figure}[htbp]
	\begin{center}
		\includegraphics[width=0.9\linewidth]{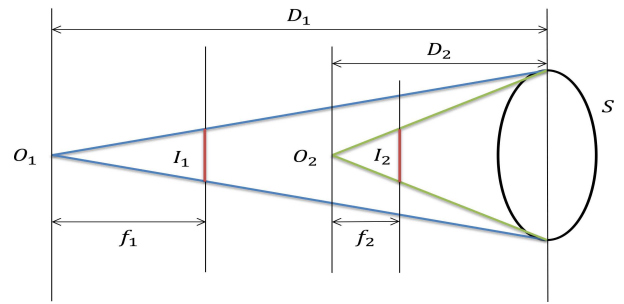}
	\end{center}
	\caption{Indistinguishability between the scene depth and the focal length in images.}
	\label{fig:ambiguity_a}
\end{figure}

Based on the imaging principle, the image of an object projected by a long-focal-length camera in the far distance could be exactly the same as the one captured by a short-focal-length camera at a short distance. This is called the indistinguishability between the scene depth and the focal length in images. For the sake of convenience, we assume that the imaging model of the camera is the pinhole model without loss of generality. For simplicity, assume the space object is linear, as shown in Figure~\ref{fig:ambiguity_a}. The images of the planar space object $ S $ under $ (f_{1}, O_{1}) $ and $ (f_{2}, O_{2}) $ are $ I_{1} $, $ I_{2} $ respectively, where $ I_{1}= I_{2} $. As a result, we are not able to infer the real depth without camera focal length from its projected image, since $ I_{1} = I_{2} $, $ D_{1}\not=D_{2} $, as shown in Figure~\ref{fig:ambiguity_a}.

\begin{figure}[htbp]
	\begin{center}
		\includegraphics[width=\linewidth]{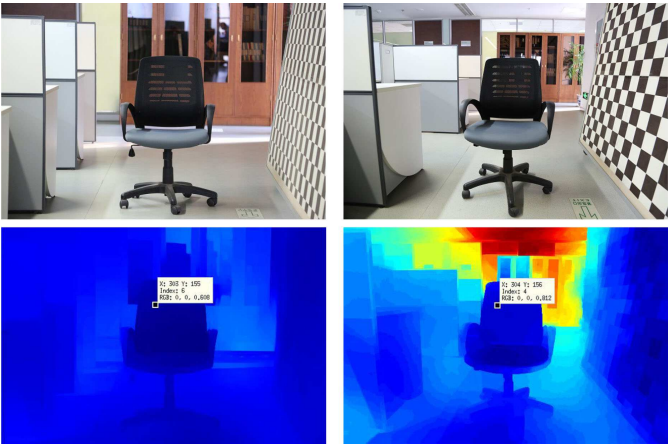}
	\end{center}
	\caption{Evaluation on depth estimation accuracy via human-computer interaction.}
	\label{fig:ambuity_human-computer}
\end{figure}

\begin{figure}[t]
\begin{center}
    \includegraphics[width=\linewidth]{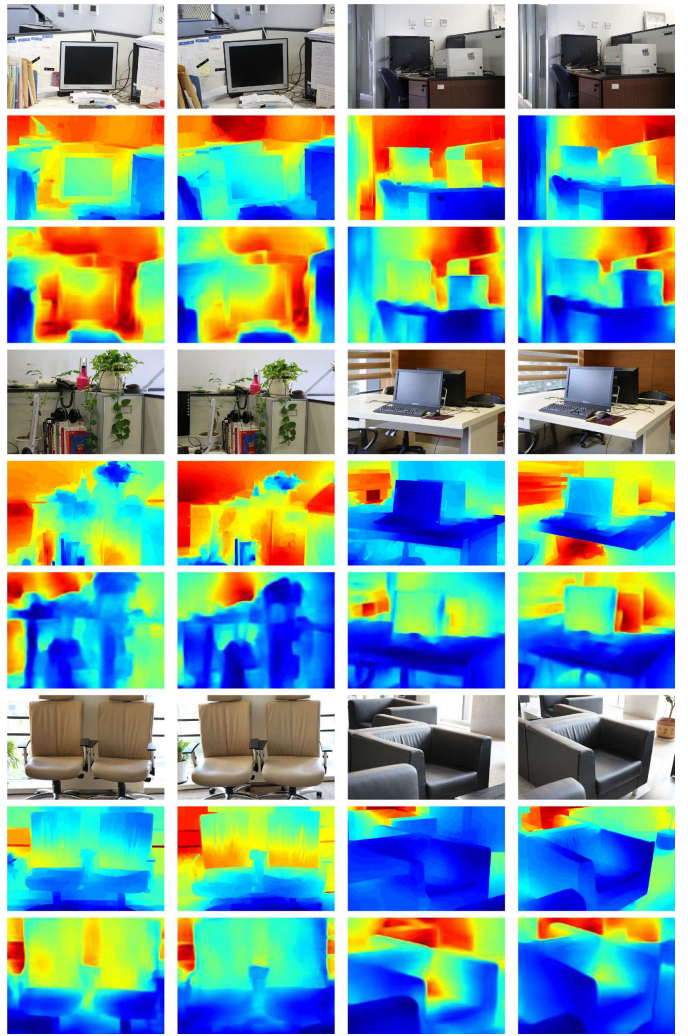}
\end{center}
   \caption{Some results of depth estimation from single images with different focal lengths. The focal lengths from left to right are 105 $ mm $ and 50 $ mm $, where first row are the RGB images, second row and third row are the inferred depths from Liu \etal~\cite{liu2015deep} and Eigen \etal~\cite{eigen2015predicting} respectively.}
\label{fig:ambuity_results}
\end{figure}

To demonstrate the ambiguity between the depth and the focal length, we collected 250 image pairs in the laboratory setting with approximate context. These images are captured by the same camera at two different focal lengths: 50 $ mm $ and 105 $ mm $, where the actual depth difference between the two images in each group is at least 1 $ m $. Then, we employ Liu~\etal~\cite{liu2015deep} and Eigen~\etal~\cite{eigen2015predicting} methods to infer the depth of the above dataset. Some experimental results are shown in Figure~\ref{fig:ambuity_results}. By human-computer interaction method, the depths of the image pairs with two focal length are measured, as shown in Figure~\ref{fig:ambuity_human-computer}. The focal length of the left image is 105 $ mm $, and the right one is 50 $ mm $. Given the depths inferred by Liu~\etal~\cite{liu2015deep}, the matching regions of the fixed object are selected to compute the average depth. The experiment shows that the average depth difference is 0.07506 $ m $, while the actual depth difference between the two images is 2 $ m $. By this measure, we take Liu~\etal~\cite{liu2015deep} and Eigen~\etal~\cite{eigen2015predicting} methods to evaluate the collected dataset,  as reported in Table~\ref{table:ambuity}, the corresponding error rates are $ 89.76\% $ and $ 87.2\% $ respectively. The experiments demonstrate that there exists inherent ambiguity between the focal length and the scene depth learned from single images.

\begin{table}[!htb]
\begin{center}
\begin{tabular}{|l|c|c|c|}
\hline
Methods & Testing pairs & Incorrect estimation pairs & Error rate \\
\hline
Liu~\etal~\cite{liu2015deep} & 250 & 224 & $ 89.6\% $ \\
Eigen \etal~\cite{eigen2015predicting} & 250 & 218 & $ 87.2\% $ \\
\hline
\end{tabular}
\end{center}
\caption{The statistical results of the depth estimation from 250 pairs of images.}
\label{table:ambuity}
\end{table}

\section{Dataset Transformation}\label{sec:data_transformation}

\begin{figure}[t]
	\begin{center}
		\includegraphics[width=0.88\linewidth]{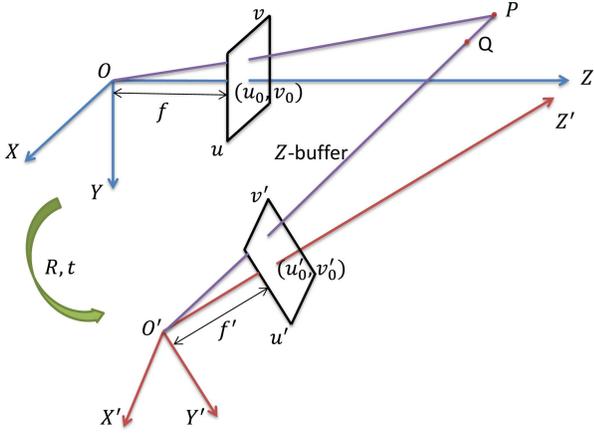}
	\end{center}
	\caption{The illustration of dataset transformation.}
	\label{fig:pipeline}
\end{figure}

In order to remove the above ambiguity, the camera's intrinsic parameter should be taken into account in the depth learning from single images, at least the focal length information should be used as input in both training and testing phases. However, all available depth datasets (like Make3D and NYU v2) in the literature are of fixed focal length. In order to remove the ambiguities caused by the focal length, we propose an approach to transform a fixed-focal-length dataset into a varying-focal-length dataset. The pipeline of the proposed approach is shown in Figure~\ref{fig:pipeline}, and the implementation details of the dataset transformation is described in the following subsections.

\subsection{Varying-focal-length image generation}
As shown in Figure~\ref{fig:pipeline}, given the camera's intrinsic parameters and the corresponding RGB-D image, the imaging process can be formulated as:

\begin{equation}\label{eq:1}
Z\left[\begin{array}{c}u\\v\\1\end{array}\right]
=\left[\begin{array}{ccc}f&0&u_{0}\\0&f&v_{0}\\0&0&1\end{array}\right]
\left[\begin{array}{c}X\\Y\\Z\end{array}\right]
\end{equation}
where $ (u_{0},v_{0}) $ is the principle point, $ f $ is the focal length, $ Z $ is the corresponding depth value, and $ (X,Y,Z) $ is the 3D space point in the camera system corresponding to the image pixel $ (u,v) $.

To transform the 3D points from the original camera coordinate to a new system, a translation and a rotation are performed according to \begin{equation}\label{eq:2}
\left[\begin{array}{c}X'\\Y'\\Z'\end{array}\right]
=R\left(\left[\begin{array}{c}X\\Y\\Z\end{array}\right]-t\right)
\end{equation}
where $ R $ is the rotation matrix, and $ t $ is the translation vector. As shown in Figure~\ref{fig:pipeline}, the camera coordinate system $(O, X, Y, Z)$ is transformed to a new system $(O', X', Y', Z')$.

By specifying a new focal length, or new camera's intrinsic matrix, the transformed 3D scene points can be projected to new image points. During the process of reprojection, multiple 3D points along the ray may be projected to the same image pixel, such as the 3D points $(P,Q)$ and pixel $(u',v')$ in Figure~\ref{fig:pipeline}. To solve this issue, we only project the 3D point with the smallest depth value, since other points are occluded by the nearest one. To obtain a real image, the new image points are quantized, and the RGB values are taken from the corresponding original image.

\begin{figure}[t]
	\begin{center}
		\includegraphics[width=\linewidth]{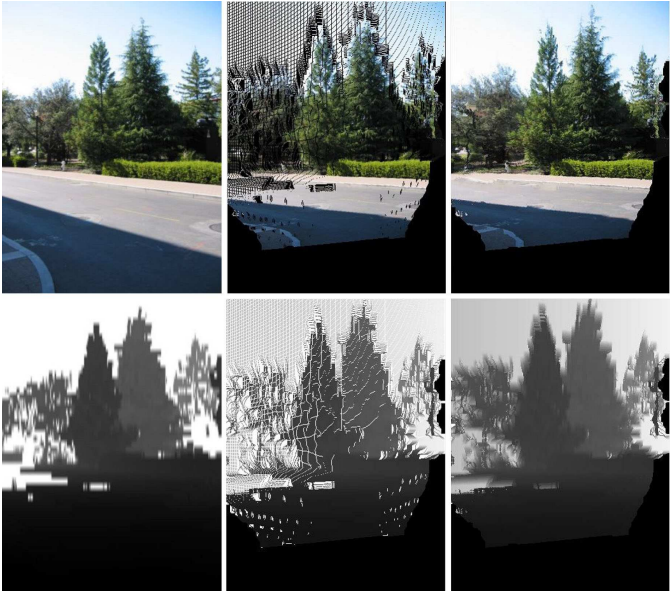}
	\end{center}
	\caption{An example with filling holes of RGB-D images, where each column represents original RGB-D images, transformed RGB-D images, and RGB-D images after postprocessing.}
	\label{fig:filling_holes}
\end{figure}

\subsection{Post-processing of the generated varying-focal-length datasets}

After the above operations, some holes are produced in the generated RGB-D image, as shown in Figure~\ref{fig:filling_holes}. By analyzing the shapes and properties of the holes, we propose a simple yet effective method to fill these holes.

\begin{figure}[t]
\begin{center}
	\includegraphics[width=1\linewidth]{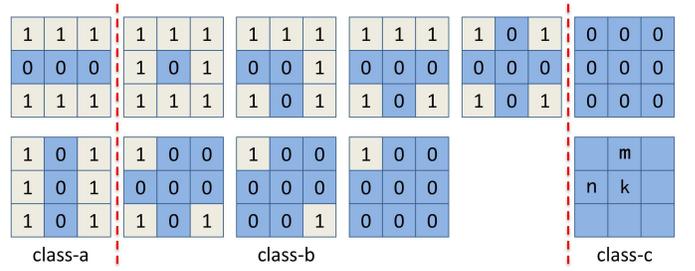}
\end{center}
    \caption{Three classes of the $3\times3$ neighborhood patterns used to fill the projected holes}
	\label{fig:filter pattern}
\end{figure}

First, we locate the positions of the empty holes, and then design $3\times3$ binary filters to fill them. The experimental holes are filled by the corresponding binary templates, which are mainly classified into three classes, as shown in Figure~\ref{fig:filter pattern}, where number 0 represents the hole pixel, and number 1 represents pixel without hole.

\begin{figure*}[htbp]
\begin{center}
    \includegraphics[width=0.94\linewidth]{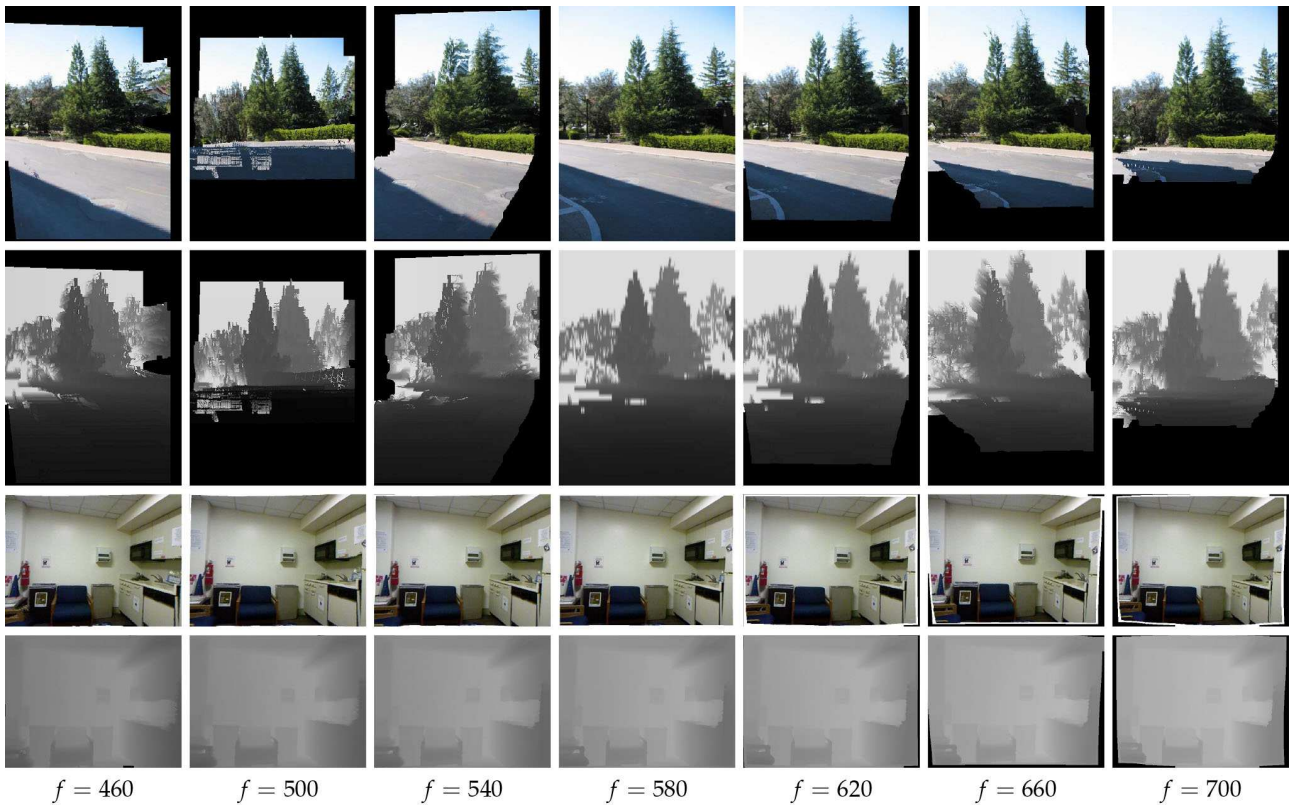}
\end{center}
   \caption{The original RGB-D image ($ f = 580 $) and sixe newly generated image sets from the Make3D dataset (top two rows) and the NYU dataset (bottom two rows).}
\label{fig:varying_focal_length_image}
\end{figure*}

For class-a, a 4-neighborhood binary template is employed for mean interpolation. For class-b, we directly use the corresponding $3\times3$ templates for mean interpolation. For class-c, the template elements all equal to zero, we iteratively perform interpolation by virtue of the intermediate interpolation results as follows: Since the iteration scheme is from left to right, and top to bottom, at least one of the two pixels at m and n has been interpolated by the previous iteration, then the (RGB-D) value at pixel k is assigned to either that at m or n with a chance.

Through the above proposed filtering process, the projected holes could be filled. Some filling results are shown in Figure~\ref{fig:filling_holes}.

\subsection{Implementation details}
Based on extensive experiments, we find that a reasonable range of the rotation angle should be within$ [-5^{\circ},5^{\circ}]$. Upon completion of the rotation, if the center of the new image coincide with the original one, the translation vector $(C_{x}, C_{y}, C_{z})$ is computed as follow.

If the rotation is around the $y$ axis by angle $\beta$, we set
\begin{equation}\label{eq:4}
\left\{
\begin{aligned}
C_{y} & =  0 \\
C_{x} & =  \frac{1}{N}(X-\frac{X+Z\sin\beta}{\cos\beta}) \\
C_{z} & =  Z-\frac{f'Z}{f\cos\beta}+(X-C_{x})\tan\beta
\end{aligned}
\right.
\end{equation}
where $N$ is the number of 3D points, and $ f' $ is the assigned new focal length.

If the rotation is around the $x$ axis by angle $\alpha$, we set
\begin{equation}\label{eq:5}
\left\{
\begin{aligned}
C_{x} & =  0 \\
C_{y} & =  \frac{1}{N}(Y-\frac{Y-Z\sin\alpha}{\cos\alpha}) \\
C_{z} & =  Z-\frac{f'Z}{f\cos\alpha}-(Y-C_{y})\tan\alpha
\end{aligned}
\right.
\end{equation}

Using the above proposed approach, we have transformed the NYU dataset and the Make3D dataset into the new varying-focal-length datasets (VFL). According to the equations~\eqref{eq:1} and~\eqref{eq:2}, the depth map of the transformed images are generated by strict geometric relationship. In the stage of quantization, some holes are introduced. However, the hole portion of the depth map is very small as shown in Figure~\ref{fig:filling_holes}, benefiting from the completion technique in equations~\eqref{eq:4} and~\eqref{eq:5}. By making use of contextual information, the holes of the depth map are filled with the proposed filtering method, which approaches to the ground truth ($f=580$) in visualization.

Figure~\ref{fig:varying_focal_length_image} shows two examples of the newly generated images from the Make3D dataset and the NYU dataset. For the generated VFL datasets, the focal-length values are 460, 500, 540, 620, 660, and 700 pixels, respectively, where the value of the initial focal length is 580. From the results we can see that the generated database is geometrically reasonable by visual verification.

\section{Learning Monocular Depth with Deep Neural Network}\label{sec:network}
In this section, based on the varying-focal-length datasets, we propose a new model to learn depth from a single image by embedding focal length information.

\subsection{Network Model}

\begin{figure*}[t]
\begin{center}
    \includegraphics[width=0.8\linewidth]{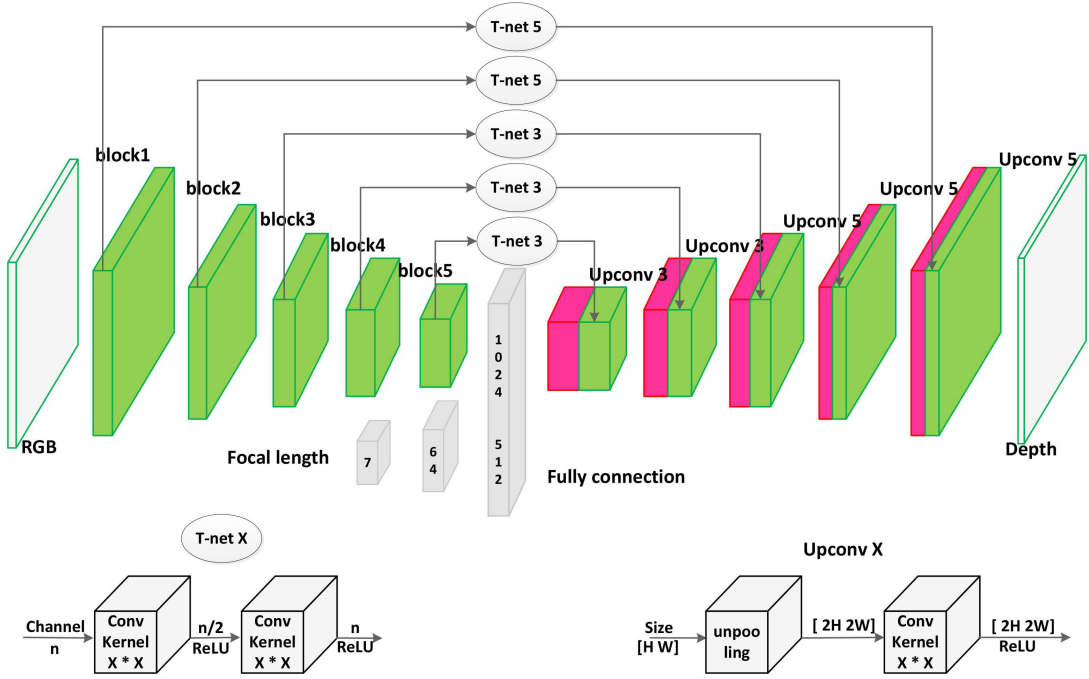}
\end{center}
   \caption{The proposed network architecture. Our neural network is built upon the pre-trained model (VGG), followed by a fully connection layer and upsampling architectures to obtain high-resolution depth, by effectively integrating the middle-level information. In addition, focal length is embedded in the network by the encoding mode.}
\label{fig:mynetwork}
\end{figure*}

The current DNN architectures are mostly built on the network~\cite{lecun1998gradient} for digit recognition, which consists of convolution, pooling, and fully connected layers. The essential power behind the remarkable success is that the framework selects the invariant abstract features which are suitable for the high-level problem. For pixel-wise depth prediction, in order to remedy the resolution loss caused by the convolution striding or pooling operations, some techniques are proposed, such as the deconvolution or upsampling methods~\cite{eigen2014depth,eigen2015predicting,laina2016deeper,xu2017multi}. Since these operations are usually applied on the last convolutional layer, it is very hard to accurately restore spacial structure information. In order to obtain pixel-wise fine-grained results, the classical skip connection is exploited, as described in the U-Net~\cite{ronneberger2015u} and the FCN~\cite{long2015fully}. For monocular depth learning, since the distribution of the depth is different from the one of the category from pre-trained model, we propose a novel transfer network (T-net), which converts feature maps from the category cues to the depth mapping, rather than utilizing feature maps directly from previous layers.

Our proposed network can be efficiently trained in an end-to-end manner,  which is symmetrical on the middle network layer, as illustrated in Figure~\ref{fig:mynetwork}. The first part of the network is based on the VGG network, which is initialized with the corresponding pre-trained weights. The second part of our architecture consists of the global transfer layer and upsampling architectures, which leads to the global information transformed from the category cues to the depth mapping and gain high resolution depth respectively. The third part of the network are T-nets, which effectively convert the middle-level information to meet the distribution of the monocular depth. The last part of our architecture are three fully connected layers for embedding the focal length information. Here, we first use the focal length to generate seven same digits, which are then connected to 64 and 512 nodes layer by layer, and finally the 512 nodes are concatenated with the global information.

For the sake of effectively fusing the middle-level information, we divide the pre-trained VGG network into 5 blocks according to the resolutions of the feature maps, as shown in the left green blocks in Figure~\ref{fig:mynetwork}. The depth of the neural networks is important for depth estimation, as described in Laina~\etal~\cite{laina2016deeper}. That means the deeper the depth of the network, the more beneficial to improving the accuracy of the depth extraction. However, very deep network may lead to a result that the actual receptive field is smaller than the theoretical receptive field, as illustrated in section~\ref{sec:ambiguity}. Inspired by this observation, we propose a fully connected layer to bridge the subsampling module and the upsampling module, which is able to obtain full-resolution receptive field and convert the global information from category to depth simultaneously. To obtain the high resolution depth, we follow the work described in~\cite{dosovitskiy2015learning} by introducing the unpooling layers, which maps each pixel into the top-left corner of a $2\times2$ (zero) kernel to double the feature map sizes, followed by a convolution implementation to fuse information, as shown in the Upconv X architecture in Figure~\ref{fig:mynetwork}.

To effectively exploit the middle layer features, we propose the T-net architecture, inspired by the ResNet~\cite{he2016deep,he2016identity} and Highway~\cite{srivastava2015highway,srivastava2015training}, to facilitate the detailed structural information propagation during both the forward and the backward stages. The identity mapping with shortcuts can facilitate the optimization of the deep network, since it iteratively generates small magnitudes of responses by passing main information layer by layer, in analogy to Taylor series. While the global information is propagated through the architecture of the first part and the second part, we utilize the T-nets to transform the detailed information in the third part. The first layer of the per T-net removes the redundant information by reducing the channels of the networks, followed by another layer to convert the feature cues. The feature maps from the T-net are concatenated with the corresponding features generated from the previous layer in the second part, followed by the unpooling and convolution operations to remedy the low resolution. As illustrated in Figure~\ref{fig:mynetwork}, the feature maps in pink color are generated from the previous layer, and the feature maps in green color are the transformed middle-level information through the T-net.

\subsection{Loss function}
The parameters of the proposed network are learned through minimizing the loss function defined on the prediction and the ground truth. In general, the mean squared error (MSE) loss is taken as the standard loss, which minimizes the squared Euclidean norm between the predictions y and the ground truth $ y^* $ :
\begin{equation}\label{eq:L2_loss}
	l_{MSE}(y - y^*) = \frac{1}{N}\sum_{y_{i}\in|N|}\Vert{y_{i} - y^*_{i}}\Vert^{2}_2
\end{equation}
where N is the number of valid pixels in the batch-size training images.

Although MSE struggles to handle the uncertain inherence in recovering lost high-frequency details, minimizing MSE encourages finding pixel-wise averages of plausible solutions, leading to blurred predictions as described in~\cite{ledig2016photo,mathieu2015deep,tatarchenko2016multi}. To solve this issue, L1 yields better detail than L2 norm. Based on our experimental study, we find that the error of depth at distant is larger than that at a close distance. Inspired by the observation, a weighted loss function is introduced by penalizing the pixels with large errors. We propagate large gradients in the locations of large errors during the training phase, which coincide with the gradient propagation of the L2 norm. As described in Zwald and Lambert-Lacroix~\cite{zwald2012berhu}, the BerHu loss function is appropriate for the above phenomena, which consists of L2 and L1 norms. Following Laina \etal~\cite{laina2016deeper}, we take the BerHu loss as the error function as below by integrating the advantages of both the L2 norm and L1 norm, resulting in accelerated optimization and detailed structure.

\begin{equation}\label{eq:Berhu_loss}
B(y - \overline{y})=\left\{
\begin{array}{lcl}
|y - \overline{y}| & & {|y - \overline{y}| < c}\\
\frac{(y - \overline{y})^{2}+c^{2}}{2c} & & {|y - \overline{y}| > c}
\end{array} \right.
\end{equation}
where $ c = 0.05max_{i}(|y_{i} - \overline{y}_{i}|)$, and $ i $ indexes the pixels in the current batch.

\section{Experiments}\label{sec:experiments}
To demonstrate the effectiveness of the proposed deep neural network and the embedded focal length for monocular depth estimation, we carry out comprehensive experiments on four publically available datasets and the synthetic datasets generated in this paper: NYU v2~\cite{Silberman:ECCV12}, Make3D~\cite{saxena2009make3d}, KITTI~\cite{Uhrig2017THREEDV}, the varying-focal-length datasets generated from section~\ref{sec:data_transformation} and SUNRGBD~\cite{Song2015SUN}. In the following subsections, we report the details of our implementation and the evaluation results.

\subsection{Experimental setup}
{\bf Datasets.} The {\bf NYU Depth v2}~\cite{Silberman:ECCV12} consists of 464 scenes, captured using a Microsoft Kinect. Followed by the official split, the training dataset is composed of 249 scenes with the 795 pair-wise images, and the testing dataset includes 215 scenes with 654 pair-wise images. In addition, the raw dataset contains 407,024 new unlabeled frames. For data augmentation, we sample equally-spaced frames out of each raw training sequence, and further align the RGB-D pairs by virtue of the provided toolbox, resulting in approximately 4k RGB-D images.

Then, the sampled raw images and 795 pair-wise images are online augmented by Eigen~\etal~\cite{eigen2014depth}. The input images and the corresponding depths are simultaneously transformed using small scaling, color transformations and flips with a chance of 0.5, then we randomly crop the augmented images and depths down to the desired size of the network. Note that the following datasets are also online augmented by the same strategy. As a result, the magnitude of samples after data augmentation on NYU depth is about 48k, which is far less than 2M for coarse network, and 1.5M for fine network, as described in Eigen~\etal~\cite{eigen2014depth}. Due to the hardware limitation, we down-sample the original frames from size $ 640\times480 $ pixels to $ 320\times224 $ as the input to the network.

The {\bf Make3D} dataset~\cite{saxena2009make3d} contains 400 training images and 134 testing images of outdoor scenes, generated from a custom 3D laser scanner. While the depth map resolution of the ground truth is only $ 305\times55 $, not matching the corresponding original RGB images with $ 1704\times2272 $ pixels, we resize all RGB-D images to $ 345\times460 $ by preserving the aspect ratio of the original images. Due to the neural network architecture and hardware limitations, we subsample the resolution of the RGB-D images to $ 160\times224 $.

The ${\bf KITTI}$ dataset~\cite{Uhrig2017THREEDV} contains 93k depth maps with corresponding raw LiDaR scans and RGB images. Following the suggestion in Uhrig~\etal~\cite{Uhrig2017THREEDV}, the training dataset is composed of 86k pair-wise images, and the testing dataset includes 1k pair-wise images selected from the full validation split. Since the LiDAR returns no measurements to the upper part of the images, we only use the bottom two thirds of the images to produce a fixed crop size of $960\times224$. In order to reduce the load of computation, we randomly crop the images from the resolution $960\times224$ to $320\times224$ during the training stage.

The {\bf varying-focal-length (VFL)} datasets contain two datasets: VFL-NYU and VFL-Make3D, which are the varying-focal-length datasets from NYU Depth v2 and Make3D respectively, as described in section~\ref{sec:data_transformation}. For VFL-NYU, the training dataset and testing dataset of every focal length are split in the official manner. Following the above NYU data argumentation, we perform the training samples argumentation using the same manner without considering the raw unaligned frames, producing approximate 30k training pairs in total. As for VFL-Make3D dataset, we implement the same operations with the above Make3D dataset, resulting in about 17k training pairs.

The {\bf SUNRGBD} dataset~\cite{Song2015SUN} contains 10,335 RGB-D images, at a similar scale as PASCAL VOC, which is captured by four different sensors - Intel RealSense 3D Camera for tablets, Asus Xtion LIVE PRO for laptops, and Microsoft Kinect versions 1 and 2 for desktop. The dataset, although constructed of various focal lengths, it is different with the dataset generated by our VFL approach. In our approach, the varying-focal-length datasets are generated from the fixed-focal-length datasets, the images with varying focal lengths are of the same scene, while in the SUNRGBD dataset, different focal-length images correspond to different scenes. In addition, the SUNRGBD dataset contains more distortion parameters caused by the four different sensors. Following the official split, the training dataset is composed of 5285 pair-wise images, and the testing dataset includes 5050 pair-wise images. In the following experiments, we sample frames out of the source dataset, resulting in 2642 pair-wise training images and 1010 pair-wise test images.

{\bf Evaluation Metrics.}
For quantitative evaluation, we report errors obtained with the following extensively adopted error metrics.
\begin{itemize}
\item
{Average relative error:
$ {\bf rel} = \frac{1}{N}\sum_{y_{i}\in|N|}\frac{|y_{i}-y^*_{i}|}{y^*_{i}} $}
\item
{Root mean squared error:

$ {\bf rms} = \sqrt{\frac{1}{N}\sum_{y_{i}\in|N|}|y_{i}-y^*_{i}|^{2}} $}
\item
{ Average $ log_{10} $ error: $ {\bf log}_{10} = \frac{1}{N}\sum_{y_{i}\in|N|}|log_{10}(y_{i})-log_{10}(y^*_{i})| $}
\item
{ Accuracy with threshold $ t $: percentage (\%) of $ y_{i} $ subject to $ max(\frac{y^*_{i}}{y_{i}}, \frac{y_{i}}{y^*_{i}})=\delta < t(t\in[1.25, 1.25^2, 1.25^3])$}
\end{itemize}
where $ y_{i} $ is the estimated depth, $ y^*_{i} $ denotes the corresponding ground truth, and $ N $ is the total number of valid pixels in all images of the validation set.

{\bf Implementation Details.} We use TensorFlow~\cite{abadi2016tensorflow} deep learning framework to implement the proposed network, and train the network on a single NVIDIA GeForce GTX TITAN with 12GB memory. The objective function is optimized using the Adam method~\cite{kingma2014adam}. During the initialization stage, weight layers in the first part of the architecture are initialized using the corresponding model (VGG) pre-trained on the ILSVRC~\cite{russakovsky2015imagenet} dataset for image classification. The weights of newly added network are assigned by sampling a Gaussian with zero mean and 0.01 variance, and the learning rate is set at 0.0001. Finally, our model is trained with a batch size of 8 for about 20 epochs.

\begin{figure}[t]
\begin{center}
\includegraphics[width=0.92\linewidth]{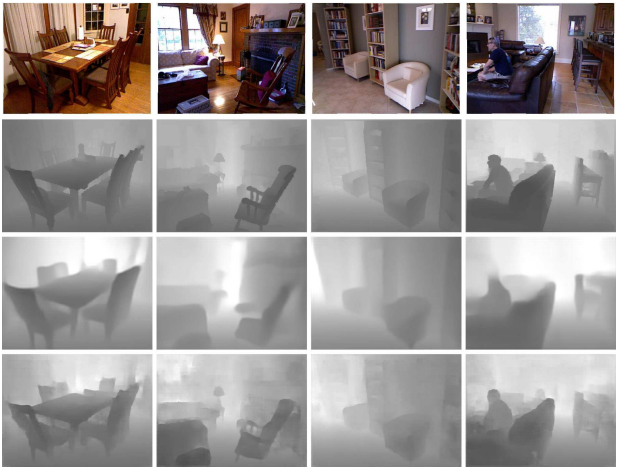}
\end{center}
\caption{Depth prediction on NYU v2 dataset. Input RGB images(first row), ground truth depth maps (second row), Laina \etal~\cite{laina2016deeper} depth results (third row) and our predictions (last row).}
\label{fig:nyu_result}
\end{figure}

\begin{table}[t]
\begin{center}
\begin{tabular}{|l|c|c|c|c|c|c|}
\hline
\multirow{2}{*}{Method} & \multicolumn{3}{c|}{Error (lower is better)} & \multicolumn{3}{c|}{Accuracy (higher is better)} \\
\cline{2-7}
& {\bf rel} & {\bf rms } & $ {\bf log}_{10} $ & $1.25$ & $1.25^2$ & $1.25^3$ \\
\hline
$ SC + L1 + G $ & 0.197 & 0.702 & 0.083 & 0.696 & 0.910 & 0.972 \\
$ T + L1 + G $ & 0.168 & 0.600 & 0.070 & 0.761 & 0.937 & 0.982 \\
$ T + B + G^{*} $ & 0.222 & 0.895 & 0.105 & 0.563 & 0.856 & 0.960 \\
$ T + B + G $ & {\bf 0.151} & {\bf 0.572} & {\bf 0.064} & {\bf 0.789} & {\bf 0.948} & {\bf 0.986} \\
\hline
\end{tabular}
\end{center}
\caption{Comparisons on the different architectures and loss functions. SC, T, L1, B, $ G^{*} $ and G represent skip connection, T-net, L1 loss, BerHu loss, GIL-convolution and GIL-connected respectively. }
\label{table:different_architecture}
\end{table}

\begin{table}[t]
\begin{center}
\begin{tabular}{|l|c|c|c|c|c|c|}
\hline
\multirow{2}{*}{Method} & \multicolumn{3}{c|}{Error (lower is better)} & \multicolumn{3}{c|}{Accuracy (higher is better)} \\
\cline{2-7}
& {\bf rel} & {\bf rms } & $ {\bf log}_{10} $ & $1.25$ & $1.25^2$ & $1.25^3$ \\
\hline
Karsch~\etal~\cite{karsch2012depth} & 0.374 & 1.12 & 0.134 & 0.447 & 0.745 & 0.897\\
Liu~\etal~\cite{liu2014discrete} & 0.335 & 1.06 & 0.127 & - & - & - \\
Li~\etal~\cite{li2015depth} & 0.232 & 0.821 & 0.094 & - & - & -\\
Liu~\etal~\cite{liu2015deep} & 0.230 & 0.824 & 0.095 & 0.614 & 0.883 & 0.975\\
Wang~\etal~\cite{wang2015towards} & 0.220 & 0.745 & 0.094 & 0.605 & 0.890 & 0.970\\
Eigen~\etal\cite{eigen2014depth} & 0.215 & 0.907 & - & 0.611 & 0.887 & 0.971\\
R. and T.~\cite{roy2016monocular} & 0.187 & 0.744 & 0.078 & - & - & -\\
E. and F.~\cite{eigen2015predicting} & 0.158 & 0.641 & - & 0.769 &{\bf 0.950} & {\bf 0.988}\\
L.-VGG~\cite{laina2016deeper} & 0.194 & 0.790 & 0.083 & - & - & -\\
E. and F. *~\cite{eigen2015predicting} & 0.155 & 0.576 & 0.065 & 0.787 & 0.948 & 0.986\\
L. *~\cite{laina2016deeper} & 0.204 & 0.833 & 0.097 & 0.617 & 0.889 & 0.963\\
ours-VGG & {\bf 0.151} & {\bf 0.572} & {\bf 0.064} & {\bf 0.789} & 0.948 & 0.986\\
\hline
\end{tabular}
\end{center}
\caption{Depth reconstruction errors on the NYU depth dataset.}
\label{table:nyu}
\end{table}

\subsection{Analysis of the different architectures and loss functions}
In the first series of experiments we focus on the NYU Depth v2~\cite{Silberman:ECCV12} dataset. The proposed model is evaluated and compared with other classical architectures and training loss functions. Specifically, we conduct the following experiments for comparison: (i) T-net and skip connection, (ii) BerHu loss and L1 loss, (ii) fully convolution (GIL-convolution) and fully connected (GIL-connected) as global information layer for bridging downsampling part and upsampling part. The results of experimental comparisons are reported in Table~\ref{table:different_architecture}. It is evident that the model with the T-net achieves better performance than the one with standard skip connection.

The table also compares the proposed model with BerHu loss and L1 loss, respectively. As expected, the model with BerHu loss yields more accurate depth. Finally, we analyze the impact of the GIL to the accuracy of the monocular depth, by comparing GIL-convolution and the GIL-connected. It is evident that the depth performance improves with the increase of the size of receptive field.

\begin{figure}[t]
\begin{center}
\includegraphics[width=0.92\linewidth]{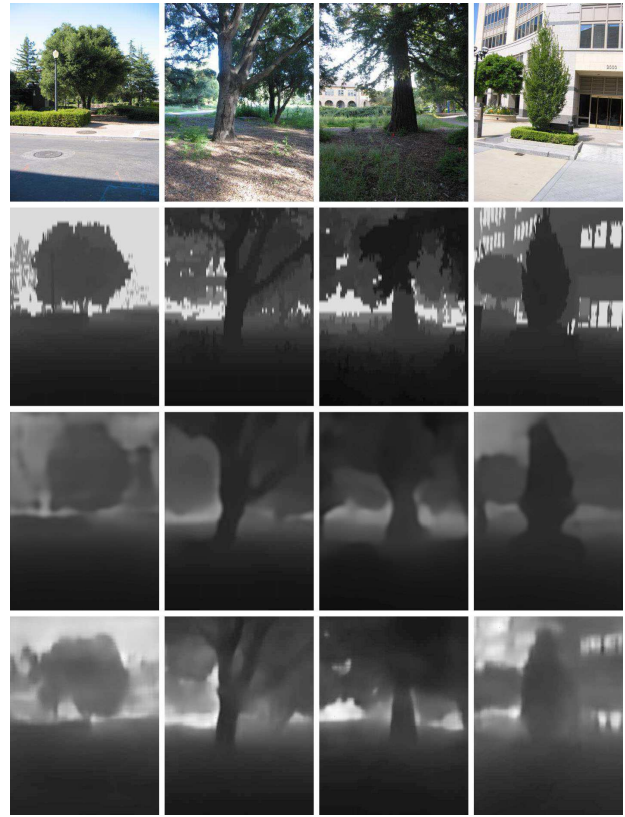}
\end{center}
\caption{Depth prediction on Make3D. Input RGB images(first row), ground truth depth maps (second row), Laina \etal~\cite{laina2016deeper} depth results (third row) and our predictions (last row). Pixels that corresponding to distances $ > 70m $ in the ground truth are masked out.}
\label{fig:make3d_result}
\end{figure}

\subsection{Comparisons with the state-of-the-art}

\begin{figure*}[t]
\begin{center}
\includegraphics[width=0.96\linewidth]{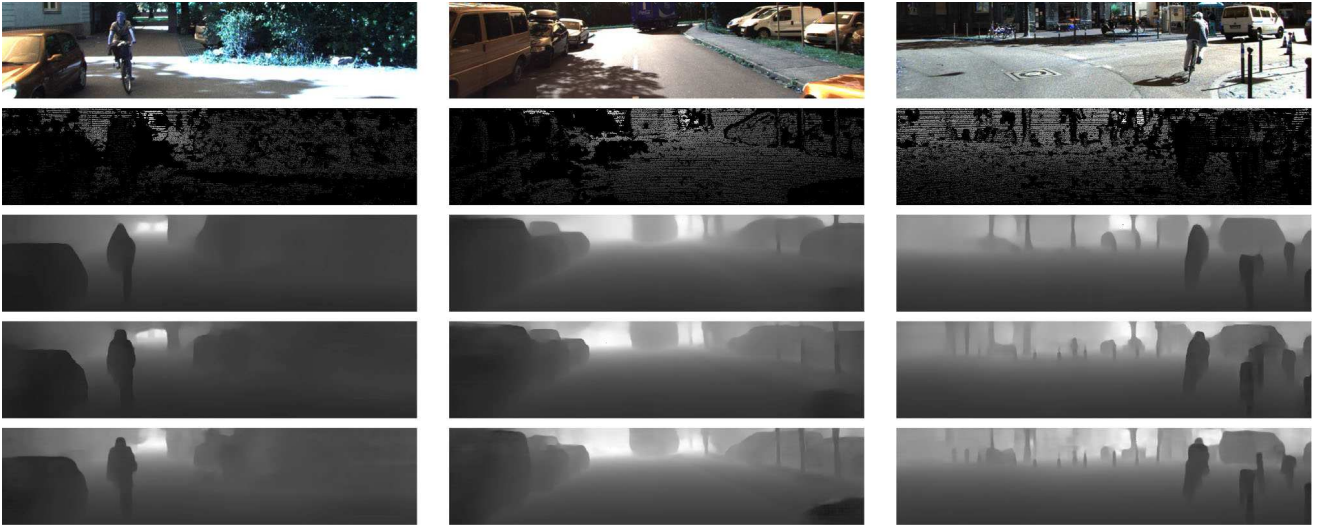}
\end{center}
\caption{Depth prediction on KITTI dataset. Input RGB images(first row), ground truth (second row), L. *~\cite{laina2016deeper} (third row), E. and F. *~\cite{eigen2015predicting} (fourth row), our proposed method (last row).}
\label{fig:kitti}
\end{figure*}

We also compared our method with the state-of-the-art approaches on NYU v2, Make3D and KITTI datasets. For the baselines, we reproduced the algorithms of VGG-Laina~\etal~\cite{laina2016deeper} and multi-scale Eigen, Fergus~\cite{eigen2015predicting} built on VGG, denoted as L. *~\cite{laina2016deeper}, and E. and F. *~\cite{eigen2015predicting}, respectively, as shown in Table~\ref{table:nyu}. For Eigen and Fergus~\cite{eigen2015predicting}, we modify the network by removing the fully connection layers in the scale 1 and directly implement upsampling operation in the last convolution layer, finally train the model in an end-to-end manner instead of the stage-wise manner. Here, the results of other algorithms are taken from the original papers. The comparative results of the proposed approach and baselines are reported in Table~\ref{table:nyu}. It is evident that our method is significantly better than the state-of-the-art approaches. By comparing VGG-Laina~\etal~\cite{laina2016deeper} with VGG-ours, we find that the effective integration of the middle-level information leads to a better performance. In addition, the performance of our reproductive algorithms is comparable with the corresponding methods. Figure~\ref{fig:nyu_result} shows some qualitative comparisons of the depth maps recovered by our method and Laina \etal~\cite{laina2016deeper} using the NYU v2 dataset. It can be seen that the estimated maps by our method can obtain more detailed information than Laina \etal~\cite{laina2016deeper}, benefiting from the effective fusion of the middle-level information with the T-net.

\begin{table}[t]
\begin{center}
\begin{tabular}{|l|c|c|c|}
\hline
\multirow{2}{*}{Method} & \multicolumn{3}{c|}{Error (lower is better)} \\
\cline{2-4}
& {\bf rel} & {\bf rms } & $ {\bf log}_{10} $ \\
\hline
Karsch~\etal~\cite{karsch2012depth} & 0.355 & 9.20 & 0.127 \\
Liu~\etal~\cite{liu2014discrete} & 0.335 & 9.49 & 0.137 \\
Liu~\etal~\cite{liu2015deep} & 0.314 & 8.60 & 0.119 \\
Li~\etal~\cite{li2015depth} & 0.278 & 7.19 & 0.092 \\
Roy and Todorovic~\cite{roy2016monocular} & 0.260 & 12.40 & 0.119 \\
E. and F. *-VGG~\cite{eigen2015predicting} & 0.228 & 7.14 & 0.093 \\
L. *-VGG~\cite{laina2016deeper} &0.236 & 7.54 & 0.091 \\
VGG-ours & {\bf 0.207} & {\bf 6.90} & {\bf 0.084} \\
\hline
\end{tabular}
\end{center}
\caption{Depth reconstruction errors on the Make3D depth dataset.}
\label{table:make3d}
\end{table}

\begin{table}[t]
\begin{center}
\begin{tabular}{|l|c|c|c|c|c|c|}
\hline
\multirow{2}{*}{Method} & \multicolumn{3}{c|}{Error (lower is better)} & \multicolumn{3}{c|}{Accuracy (higher is better)} \\
\cline{2-7}
& {\bf rel} & {\bf rms } & $ {\bf log}_{10} $ & $1.25$ & $1.25^2$ & $1.25^3$ \\
\hline
E. and F. *~\cite{eigen2015predicting} & 0.095 & 4.131 & 0.042 & 0.893 & {\bf 0.976} & 0.993\\
L. *~\cite{laina2016deeper} & 0.108 & 4.326 & 0.049 & 0.874 & 0.975 & 0.993\\
ours-VGG & {\bf 0.086} & {\bf 4.014} & {\bf 0.040} & 0.893 & 0.975 & {\bf 0.994}\\
\hline
\end{tabular}
\end{center}
\caption{Depth reconstruction errors on the KITTI depth dataset.}
\label{table:kitti}
\end{table}

\begin{figure}[t]
\begin{center}
\includegraphics[width=0.98\linewidth]{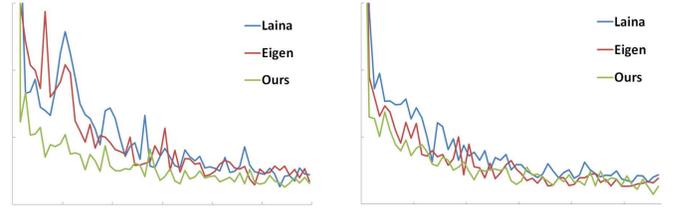}
\end{center}
\caption{Training error on the KITTI dataset (left) and the NYU v2 dataset (right).}
\label{fig:trainin_error}
\end{figure}

\begin{table}[!tb]
\begin{center}
\begin{tabular}{|l|c|c|c|}
\hline
\multirow{2}{*}{Method} & NYU & Make3D & KITTI \\
&($640\times480$) & ($345\times460$) & ($960\times224$) \\
\hline
E. and F. *~\cite{eigen2015predicting} & 0.269 & 0.137 & 0.194 \\
L. *~\cite{laina2016deeper} & {\bf 0.182} & {\bf 0.098} & {\bf 0.142} \\
ours-VGG & 0.202 & 0.101 & 0.150 \\
\hline
\end{tabular}
\end{center}
\caption{Execution time (seconds) of the proposed algorithm and the state-of-the-art approaches on the public datasets.}
\label{table:execution_time}
\end{table}

In addition, we evaluated our proposed model on the Make3D dataset~\cite{saxena2009make3d}, generated from a custom 3D laser scanner. Following~\cite{eigen2015predicting,laina2016deeper}, the error metrics are computed on the regions with ground truth depth maps less than 70m. We also reproduce the algorithms of VGG-Laina~\etal~\cite{laina2016deeper} and multi-scale Eigen and Fergus~\cite{eigen2015predicting} with VGG as L. *-VGG~\cite{laina2016deeper} and E. and F. *-VGG~\cite{eigen2015predicting} in Table~\ref{table:make3d}. Our modified E. and F. *-VGG~\cite{eigen2015predicting} and VGG-ours outperform other methods by a significant margin, which reveals that the middle-level information is useful for the accurate depth inference, as well as multi-scale information. As expected, our proposed method yields more detailed structural information of the depth compared with Laina \etal~\cite{laina2016deeper}, as shown in Figure~\ref{fig:make3d_result}.

\begin{figure*}[t]
\begin{center}
\includegraphics[width=0.9\linewidth]{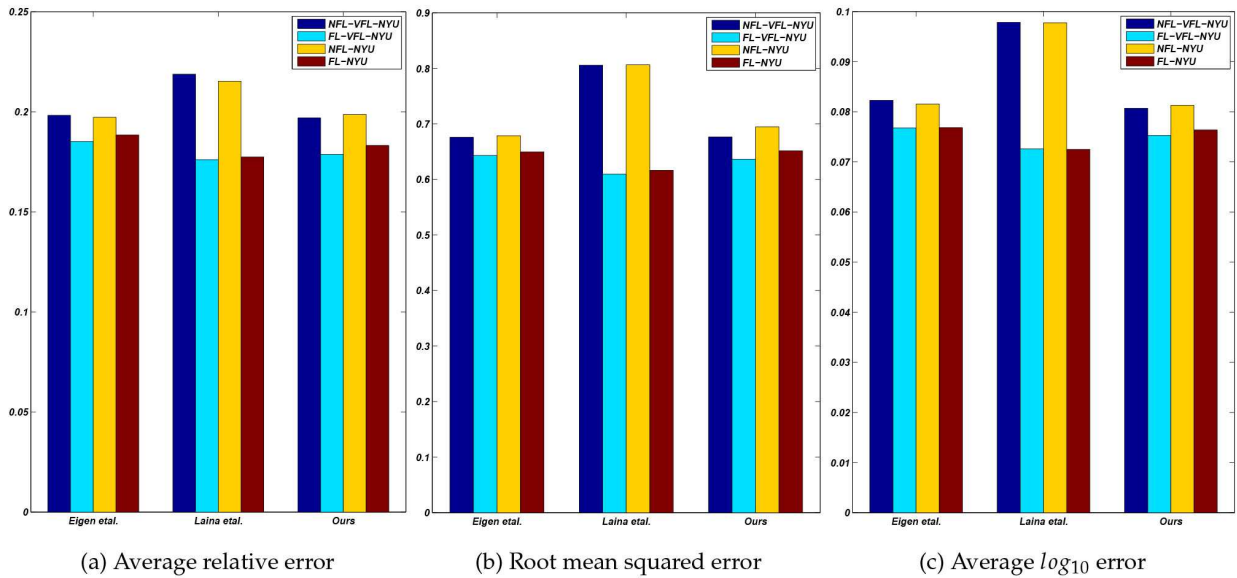}
\end{center}
\caption{Depth reconstruction errors on the VFL-NYU test dataset and NYU test dataset.}
\label{fig:result_vfl_nyu}
\end{figure*}

\begin{table*}[!htb]
\begin{center}
\begin{tabular}{|l|c|c|c|c|c|c|c|c|c|c|c|c|}
\hline
\multirow{3}{*}{Method} & \multicolumn{6}{c|}{VFL-NYU test set ($654\times7$)} & \multicolumn{6}{c|}{NYU test set (654)}\\
\cline{2-13}
& \multicolumn{3}{c|}{Error (lower is better)} & \multicolumn{3}{c|}{Accuracy (higher is better)} & \multicolumn{3}{c|}{Error (lower is better)} & \multicolumn{3}{c|}{Accuracy (higher is better)}  \\
\cline{2-13}
& {\bf rel} & {\bf rms } & $ {\bf log}_{10} $ & $\delta<1.25$ & $\delta<1.25^2$ & $\delta<1.25^3$ & {\bf rel} & {\bf rms } & $ {\bf log}_{10} $ & $\delta<1.25$ & $\delta<1.25^2$ & $\delta<1.25^3$\\
\hline
\cite{eigen2015predicting}-NFL & 0.198 & 0.676 & 0.082 & 0.659 & 0.882 & 0.938 & 0.197 & 0.679 & 0.082 & 0.693 & 0.919 & 0.978\\
\cite{eigen2015predicting}-FL & {\bf 0.186} & {\bf 0.643} & {\bf 0.077} & {\bf 0.693} & {\bf 0.889} & {\bf 0.939} & {\bf 0.188} & {\bf 0.650} & {\bf 0.077} & {\bf 0.721} & {\bf 0.923} & 0.978\\
\hline
\cite{laina2016deeper}-NFL & 0.219 & 0.806 & 0.098 & 0.566 & 0.847 & 0.932 & 0.215 & 0.806 & 0.098 & 0.584 & 0.884 & 0.974\\
\cite{laina2016deeper}-FL & {\bf 0.176} & {\bf 0.609} & {\bf 0.073} & {\bf 0.716} & {\bf 0.899} & {\bf 0.942} & {\bf 0.177} & {\bf 0.617} & {\bf 0.073} & {\bf 0.746} & {\bf 0.935} & {\bf 0.982}\\
\hline
ours-NFL & 0.197 & 0.677 & 0.081 & 0.668 & 0.884 & 0.939 & 0.199 & 0.694 & 0.081 & 0.694 & 0.916 & 0.976\\
ours-FL & {\bf 0.177} & {\bf 0.636} & {\bf 0.075} & {\bf 0.694} & {\bf 0.895} & {\bf 0.944} & {\bf 0.183} & {\bf 0.651} & {\bf 0.076} & {\bf 0.715} & {\bf 0.928} & {\bf 0.983}\\
\hline
\end{tabular}
\end{center}
\caption{Depth reconstruction errors on the VFL-NYU dataset and NYU dataset.}
\label{table:nyu_w/o_focal_length}
\end{table*}

Furthermore, considering that the Make3D~\cite{saxena2009make3d} is a very small dataset, to prove the advantage of the proposed model in the outdoor images, we further evaluate the proposed approach on the KITTI dataset~\cite{Uhrig2017THREEDV}. Due to the resolution difference of the training images and the testing images, we replace the fully connected layer of our proposed network with $1\times1$ fully convolution layer. To achieve a fair comparison with the state-of-the-art methods, we also reproduce the algorithms of L. *-VGG~\cite{laina2016deeper} and E. and F. *-VGG~\cite{eigen2015predicting} as above. The quantitative results of each approach are reported in Table~\ref{table:kitti}. It is clear that the proposed approach yields lower error than both the L. *-VGG~\cite{laina2016deeper} and the L. *-VGG~\cite{laina2016deeper} approachs, which demonstrates the advantage of the proposed model. As shown in Figure~\ref{fig:kitti}, compared with L. *-VGG~\etal~\cite{laina2016deeper} and E. and F. *-VGG~\etal~\cite{eigen2015predicting}, two of the best methods in the literature, it is evident that our approach achieves better fine-grained depth in visualization. Note that our method and the reproduced algorithms utilize sparse point information to infer dense depth from a single image, which reveals that these methods can also be used in 3D LiDARs to address depth completion problem.

In addition, we also compared the execution time between the proposed method and the state-of-the-art algorithms. Table~\ref{table:execution_time} tabulates the real runtime on the NYU v2, Make3D, and KITTI datasets, corresponding to resolution of $640\times480$, $345\times460$ and $960\times224$, respectively. L. *~\cite{laina2016deeper} is the fastest algorithm since it has less number of convolutional layers and filters. Since the proposed method exploits T-nets to fuse middle-level information, it runs a little bit slower than the L. *~\cite{laina2016deeper} algorithm. However, the speed of our approach still performs favorably against the E. and F. *~\cite{eigen2015predicting} algorithm as the later one utilizes large convolutional kernel to integrate multi-scale information. It is worth noting that it only takes about 0.1 sec in total for our method to recovery the depth map for a single image ($320\times224$), which enables the possibility of inferring fine-grained monocular depth in real-time.

To evaluate the convergence process of the proposed method, the training curves of the NYU v2 dataset and the Make3D database are visualized in Figure~\ref{fig:trainin_error}, and the state-of-the-art approaches are also implemented for comparison. It is notable that our algorithm exhibits lower training error, especially for the KITTI dataset, which contributes the performance gains in Table~\ref{table:nyu} and Table~\ref{table:kitti}. In addition, our proposed method converges faster than the L. *-VGG~\cite{laina2016deeper} and the E. and F. *-VGG~\cite{eigen2015predicting}, which facilitates the optimization by providing faster convergence at the early stage, benefiting from the T-net architecture. These comparisons verify the effectiveness of the proposed method for learning depth from a single image.

\begin{figure*}[t]
\begin{center}
\includegraphics[width=0.9\linewidth]{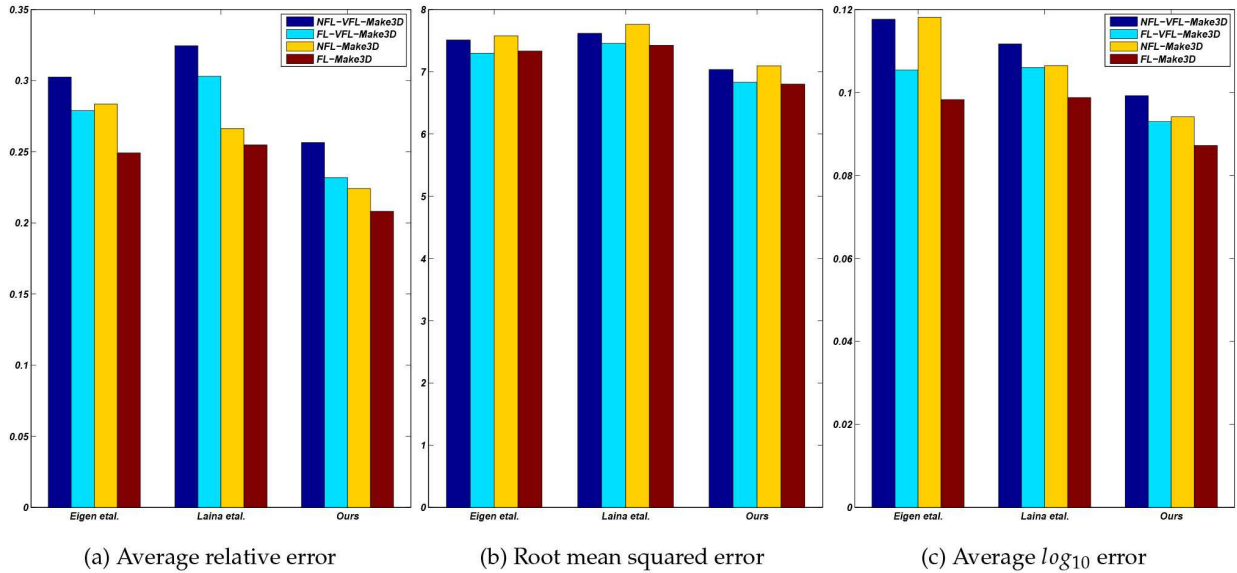}
\end{center}
\caption{Depth reconstruction errors on the VFL-Make3D test dataset and the Make3D test dataset.}
\label{fig:result_vfl_make}
\end{figure*}

\begin{table*}[!htb]
\begin{center}
\begin{tabular}{|l|c|c|c|c|c|c|c|c|c|c|c|c|}
\hline
\multirow{3}{*}{Method} & \multicolumn{6}{c|}{VFL-Make3D test set ($134\times7$)} & \multicolumn{6}{c|}{Make3D test set (134)}\\
\cline{2-13}
& \multicolumn{3}{c|}{Error (lower is better)} & \multicolumn{3}{c|}{Accuracy (higher is better)} & \multicolumn{3}{c|}{Error (lower is better)} & \multicolumn{3}{c|}{Accuracy (higher is better)}  \\
\cline{2-13}
& {\bf rel} & {\bf rms } & $ {\bf log}_{10} $ & $\delta<1.25$ & $\delta<1.25^2$ & $\delta<1.25^3$ & {\bf rel} & {\bf rms } & $ {\bf log}_{10} $ & $\delta<1.25$ & $\delta<1.25^2$ & $\delta<1.25^3$\\
\hline
\cite{eigen2015predicting}-NFL & 0.303 & 7.510 & 0.118 & 0.472 & 0.695 & 0.775 & 0.283 & 7.577 & 0.118 & 0.521 & 0.809 & 0.908\\
\cite{eigen2015predicting}-FL & {\bf 0.279} & {\bf 7.298} & {\bf 0.105} & {\bf 0.527} & {\bf 0.710} & {\bf 0.778} & {\bf 0.249} & {\bf 7.332} & {\bf 0.098} & {\bf 0.620} & {\bf 0.839} & {\bf 0.914}\\
\hline
\cite{laina2016deeper}-NFL & 0.325 & 7.616 & 0.112 & 0.505 & 0.701 & 0.771 & 0.266 & 7.762 & 0.106 & 0.595 & 0.827 & 0.904\\
\cite{laina2016deeper}-FL & {\bf 0.303} & {\bf 7.456} & {\bf 0.106} & {\bf 0.518} & {\bf 0.704} & {\bf 0.773} & {\bf 0.255} & {\bf 7.423} & {\bf 0.099} & {\bf 0.617} & {\bf 0.829} & {\bf 0.911}\\
\hline
ours-NFL & 0.256 & 7.035 & 0.099 & 0.516 & 0.676 & 0.760 & 0.224 & 7.095 & 0.094 & 0.608 & 0.784 & 0.881\\
ours-FL & {\bf 0.232} & {\bf 6.830} & {\bf 0.093} & {\bf 0.539} & {\bf 0.683} & {\bf 0.769} & {\bf 0.208} & {\bf 6.801} & {\bf 0.087} & {\bf 0.641} & {\bf 0.794} & {\bf 0.894}\\
\hline
\end{tabular}
\end{center}
\caption{Depth reconstruction errors on the VFL-Make3D dataset and Make3D dataset.}
\label{table:make3d_w/o_focal_length}
\end{table*}

\subsection{Evaluations of VFL dataset with focal length information}
Given varying-focal-length datasets generated in section~\ref{sec:data_transformation}, we utilize the network of the section~\ref{sec:network} to learn the depth from a single image, where the focal length is embedded in the network during the phases of training and testing. For comparison, the experiments are also implemented on L. *-VGG~\cite{laina2016deeper} and E. and F. *-VGG~\cite{eigen2015predicting} respectively. For E. and F. *-VGG~\cite{eigen2015predicting}, the focal length information is embedded in the last convolutional layer of the scale 1, as similar with the section~\ref{sec:network}. We implement the same operation on the last layer of the downsampling part in the L. *-VGG~\cite{laina2016deeper}. In addition, the experiments without focal length are also implemented on the above models for comparison.

For VFL-NYU dataset, the experimental results are reported in Table~\ref{table:nyu_w/o_focal_length}, where NFL denotes the model without embedded focal length, FL denotes the model with embedded focal length. At the same time, the learned models from VFL-NYU dataset are also implemented on the NYU test dataset. As shown in the Table, for average relative error,~\cite{eigen2015predicting}-FL, ours-FL and~\cite{laina2016deeper}-FL increase the accuracy by about two percentage points on average, compared with corresponding methods without embedded focal length information. Figure~\ref{fig:result_vfl_nyu} shows the increase of accuracy in the form of histogram, which reveals that each model with embedded focal length obtains much better performance than that without the focal length, where L. *-VGG~\cite{laina2016deeper} achieves a significant margin, benefiting from that the network with only one path could effectively deliver the focal length information during forward and backward phases.

We also implement our approach and the state-of-the-art methods~\cite{laina2016deeper,eigen2015predicting} on the VFL-Make3D dataset, as reported in Table~\ref{table:make3d_w/o_focal_length}, where the same trained model is also implemented on the Make3D test dataset. It is evident that, for average relative error, the three approaches with embedded focal length information also increase the accuracy by about two percentage points on average, compared with the corresponding methods without the focal length information. As shown in Figure~\ref{fig:result_vfl_make}, all models with the embedded focal length information outperform the corresponding models without the focal length information. However, the performance gains of the VFL-Make3D dataset on root square error is not as good as that of the VFL-NYU dataset, which is caused by the accuracy range of the ground truth and the training dataset size.

From Table~\ref{table:nyu_w/o_focal_length} and Table~\ref{table:make3d_w/o_focal_length}, it is notable that the models trained on the VFL-NYU dataset and VFL-Make3D dataset achieve better performance than the corresponding models without the embedded focal length information on the NYU test dataset and Make3D test dataset, which also reveals that the focal length information contributes to the performance increase in depth estimation from single images. However, compared the Table~\ref{table:nyu_w/o_focal_length} with Table~\ref{table:nyu}, the experimental results of the nets on the VFL-NYU dataset show slight weakness than the corresponding ones trained on the NYU depth. This phenomena is mainly caused by the fact that the VFL-NYU dataset is much less than the NYU dataset with raw video frames. For the model trained on the NYU depth, except for the 795 pair-wise images, we also fetch 4,000 samples from the raw dataset by virtue of the provided toolbox. While the VFL-NYU dataset is only generated from 1,449 pair-wise images, which has less samples than the models in Table~\ref{table:nyu}. In addition, The VFL-Make3D and Make3D database have approximately same number of samples, which achieve lower error difference than the VFL-NYU and the NYU datasets, as reported in Table~\ref{table:make3d_w/o_focal_length} and Table~\ref{table:make3d}.

To further prove the benefits of embedding focal length, we also performed experiments on the SUNRGBD~\cite{Song2015SUN} dataset. In order to achieve a fair comparison with the state-of-the-art methods, we also reproduce the algorithms of E. and F. *-VGG~\cite{eigen2015predicting} and L. *-VGG~\cite{laina2016deeper} in the same way. The quantitative results of each approach are reported in Table~\ref{table:sunrgbd}. The experimental results show that depths inferred from the model with embedded focal length significantly outperform those without the focal length information in all error measures, which demonstrates the contribution of the focal length information for depth estimation from a single image.

\begin{table}[t]
\begin{center}
\begin{tabular}{|l|c|c|c|c|c|c|}
\hline
\multirow{2}{*}{Method} & \multicolumn{3}{c|}{Error (lower is better)} & \multicolumn{3}{c|}{Accuracy (higher is better)} \\
\cline{2-7}
& {\bf rel} & {\bf rms } & $ {\bf log}_{10} $ & $1.25$ & $1.25^2$ & $1.25^3$ \\
\hline
\cite{eigen2015predicting}-NFL & 0.318 & 0.806 & 0.149 & 0.387 & 0.753 & 0.904\\
\cite{eigen2015predicting}-FL & {\bf 0.278} & {\bf 0.677} & {\bf 0.117} & {\bf 0.606} & {\bf 0.853} & {\bf 0.923}\\
\hline
\cite{laina2016deeper}-NFL & 0.325 & 0.834 & 0.161 & 0.419 & 0.743 & 0.874\\
\cite{laina2016deeper}-FL & {\bf 0.288} & {\bf 0.577} & {\bf 0.095} & {\bf 0.684} & {\bf 0.886} & {\bf 0.949}\\
\hline
ours-NFL & 0.294 & 0.736 & 0.139 & 0.585 & 0.822 & 0.899\\
ours-FL & {\bf 0.274} & {\bf 0.700} & {\bf 0.120} & {\bf 0.598} & {\bf 0.859} & {\bf 0.938}\\
\hline
\end{tabular}
\end{center}
\caption{Depth reconstruction errors on the SUNRGBD depth dataset.}
\label{table:sunrgbd}
\end{table}

The above experiments demonstrate that we can boost the inference accuracy of the depth when the focal length is embedded in the network in both learning and inference phases.

\section{Conclusion}\label{sec:conclusion}
In this paper, focusing on the monocular depth learning problem, we first studied the inherent ambiguity between the scene depth and the focal length in theory, and verified it using real images. In order to remove the ambiguity, we proposed an approach to generate the varying-focal-length datasets from the public fixed-focal-length datasets. Then, a novel deep neural network was proposed to infer the fine-grained monocular depth from both the fixed- and varying-focal-length datasets. We demonstrated that the proposed model, without the embedded focal length information, could achieve competitive performance on the public datasets with the state-of-the-art methods. Furthermore, by using the newly generated and publicly varying-focal-length datasets, the proposed approach and the state-of-the-art algorithms embedding focal length yield a significant performance increase in all error metrics, compared with the corresponding models without encoding focal length. The extensive experiments demonstrate that the embedding focal length is able to improve the depth learning accuracy from single images.

\section*{Acknowledgement}
This work was supported by National Natural Science Foundation of China under the grant No (61333015, 61421004, 61772444, 61573351).

\bibliographystyle{IEEEtran}
\bibliography{Learning Depth from Single Images with Deep Neural Network Embedding Focal Length}

\begin{biography}[{\includegraphics[width=1in,clip,keepaspectratio]{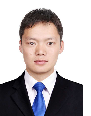}}]{Lei He} obtained his Bachelor's degree from Beijing University of Aeronautics and Astronautics, China. He is currently a PhD candidate at the National Laboratory of Pattern Recognition, Chinese Academy of Sciences. His research interests include computer vision, machine learning, and pattern recognition.
\end{biography}
\begin{biography}[{\includegraphics[width=1in,clip,keepaspectratio]{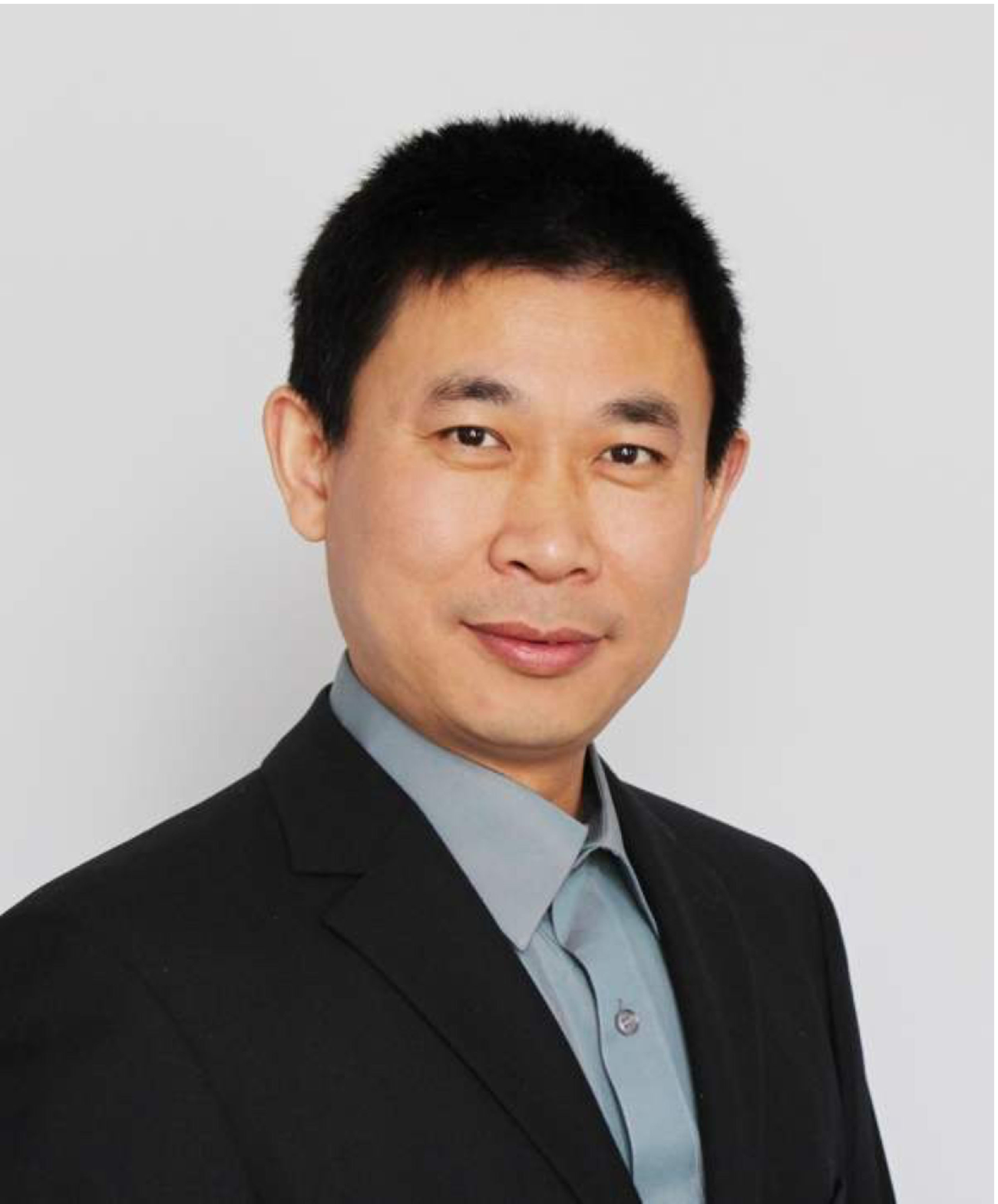}}]{Guanghui Wang} (M' 10, SM' 17) is currently an assistant professor at the University of Kansas, USA. He is also with the Institute of Automation, Chinese Academy of Sciences, China, as an adjunct professor. From 2003 to 2005, he was a research fellow and visiting scholar with the Department of Electronic Engineering at the Chinese University of Hong Kong. From 2006 to 2010, He was a research fellow with the Department of Electrical and Computer Engineering, University of Windsor, Canada. He has authored one book, "Guide to Three Dimensional Structure and Motion Factorization", published at Springer-Verlag. He has published over 100 papers in peer-reviewed journals and conferences. His research interests include computer vision, structure from motion, object detection and tracking, artificial intelligence, and robot localization and navigation. Dr. Wang has served as associate editor and on the editorial board of two journals, as an area chair or TPC member of 20+ conferences, and as a reviewer of 20+ journals.
\end{biography}
\begin{biography}[{\includegraphics[width=1in,clip,keepaspectratio]{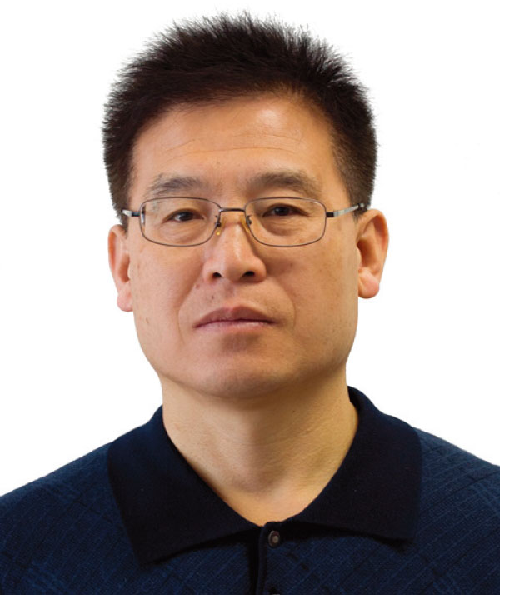}}]{Zhanyi Hu} received his B.S. in Automation from the North China University of Technology, Beijing, China, in 1985, and the Ph.D. in Computer Vision from the University of Liege, Belgium, in 1993. Since 1993, he has been with the National Laboratory of Pattern Recognition at Institute of Automation, Chinese Academy of Sciences, where he is now a professor. His research interests are in robot vision, which include camera calibration and 3D reconstruction, vision guided robot navigation. He was the local chair of ICCV 2005, an area chair of ACCV 2009, and the PC chair of ACCV 2012.
\end{biography}

\end{document}